\pdfoutput=1
\documentclass[10pt,journal,compsoc]{IEEEtran}

\ifCLASSOPTIONcompsoc
  \usepackage[nocompress]{cite}
\else
  \usepackage{cite}
\fi
\ifCLASSINFOpdf
  \usepackage[pdftex]{graphicx}
\else
\fi
\usepackage{booktabs}
\usepackage{multirow}
\usepackage{amsmath}
\usepackage{amssymb}
\usepackage{xcolor}

\def\sub#1{_{\rm #1}}
\def\vct#1{\mbox{\boldmath $#1$}}
\def\eg{{\it e.g.}}
\def\ie{{\it i.e.}}
\def\etal{{\it et al.}}
\def\etc{{\it etc.}}
\begin{document}
%
\title{Ego-Surfing: Person Localization in First-Person Videos Using Ego-Motion Signatures}
%
%
%
%

\author{Ryo~Yonetani,~\IEEEmembership{Member,~IEEE,}
        Kris~M.~Kitani,~\IEEEmembership{Member,~IEEE,}
        and~Yoichi~Sato,~\IEEEmembership{Member,~IEEE}
\IEEEcompsocitemizethanks{\IEEEcompsocthanksitem R. Yonetani and Y. Sato are with the Institute of Industrial Science, the University of Tokyo,
Japan.\protect\\
E-mail: \{yonetani, ysato\}@iis.u-tokyo.ac.jp
\IEEEcompsocthanksitem K. Kitani is with the Robotics Institute, Computer Vision Group at 
Carnegie Mellon University, USA.\protect\\
E-mail: kkitani@cs.cmu.edu}
}

%
%

\markboth{IEEE TRANSACTIONS ON PATTERN ANALYSIS AND MACHINE INTELLIGENCE}%
{Yonetani \MakeLowercase{\textit{et al.}}: EGO-SURFING FIRST-PERSON VIDEOS}
%



\IEEEtitleabstractindextext{%
\begin{abstract}
We envision a future time when wearable cameras are worn by the masses and recording first-person point-of-view videos of everyday life. While these cameras can enable new assistive technologies and novel research challenges, they also raise serious privacy concerns. For example, first-person videos passively recorded by wearable cameras will necessarily include anyone who comes into the view of a camera -- with or without consent. Motivated by these benefits and risks, we developed a self-search technique tailored to first-person videos. The key observation of our work is that the egocentric head motion of a target person (\ie, the self) is observed both in the point-of-view video of the target and observer. The motion correlation between the target person's video and the observer's video can then be used to identify instances of the self uniquely. We incorporate this feature into the proposed approach that computes the motion correlation over densely-sampled trajectories to search for a target individual in observer videos. Our approach significantly improves self-search performance over several well-known face detectors and recognizers. Furthermore, we show how our approach can enable several practical applications such as privacy filtering, target video retrieval, and social group clustering.
\end{abstract}

}

\maketitle

\IEEEdisplaynontitleabstractindextext

%
\IEEEpeerreviewmaketitle

\ifCLASSOPTIONcompsoc
\IEEEraisesectionheading{\section{Introduction}\label{sec:introduction}}
\else
\section{Introduction}
\label{sec:introduction}
\fi
\IEEEPARstart{N}{ew} technologies for image acquisition, such as wearable eye glass cameras or lapel cameras, can enable new assistive technologies and novel research challenges but may also come with latent social consequences. Around the world hundreds of millions of camera-equipped mobile phones can be used to capture special moments in life. Novel wearable camera technologies (\eg, the Google Glass or the Narrative lapel camera) also offer a new paradigm for keeping a visual record of everyday life in the form of \emph{first-person point-of-view videos} and can be used to aid productivity, such as automatic activity summarization~\cite{Arev2014, Lee2012, Lu2013, Xu2015} and assistive systems~\cite{Leung2014, Tang2014, Tian2013}.

However, as a mobile phone capturing an image with a GPS position embedded EXIF tag could be used as a means of violating one's privacy, wearable cameras also hold the inevitable risk of unintended use. Namely, first-person videos passively recorded by the wearable cameras will necessarily include anyone who comes into the view of a camera -- with or without consent. Without proper mechanisms and technologies to preserve privacy, wearable cameras run the risk of inadvertently capturing sensitive information.

\begin{figure}[t]
\centering
\includegraphics[width=\linewidth]{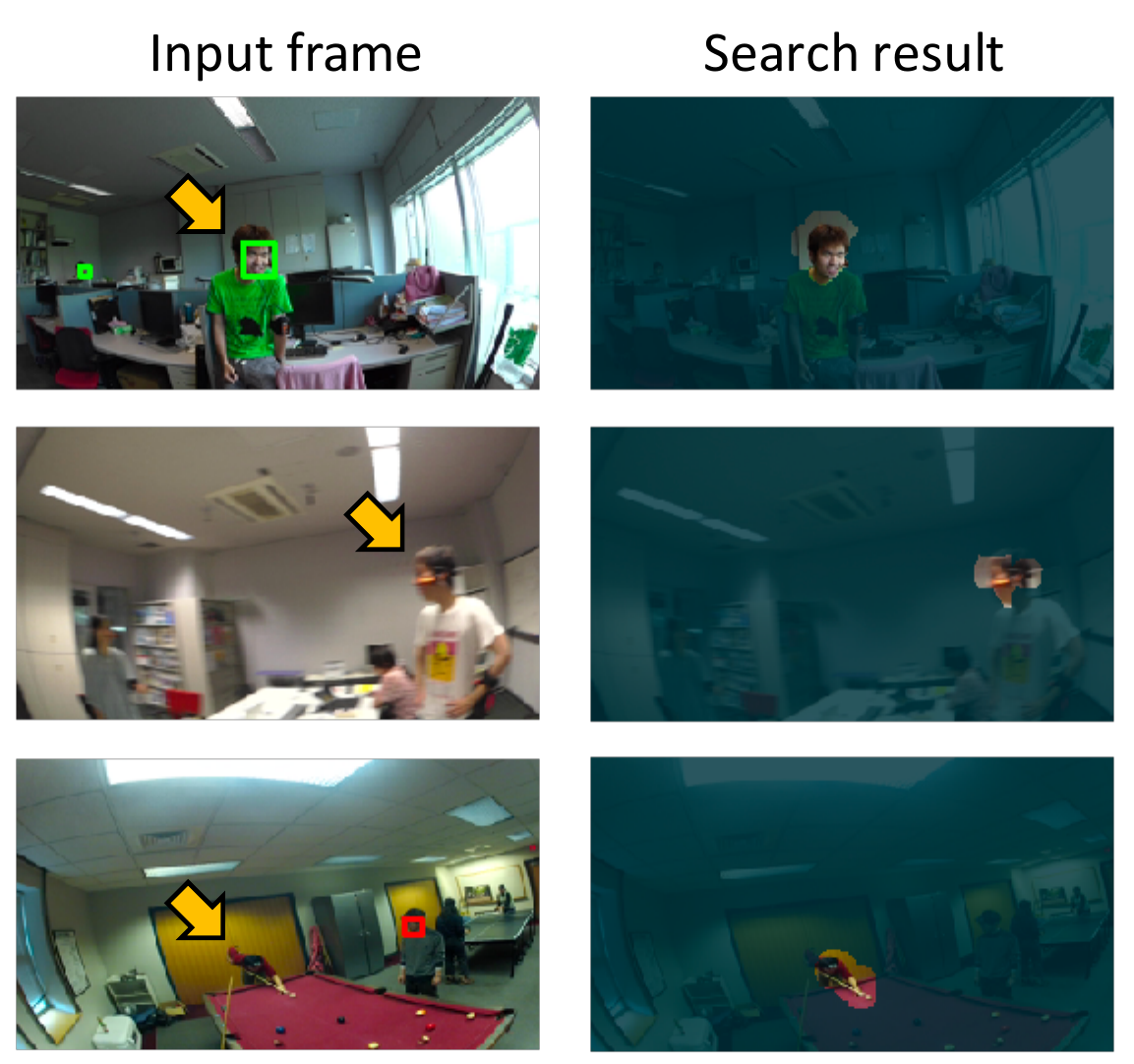}
\caption{Self-search results. Target instances ({\bf yellow arrows}) are detected ({\bf unmasked regions}) despite motion blurred, non-frontal faces ({\bf second column}) or extreme body poses ({\bf third column}) where face detection ({\bf green rectangles}) and recognition ({\bf red rectangles}) fail.}
\label{fig:teaser}
\end{figure}

Keeping in mind both the benefits and risks of using wearable cameras, we argue that one important technology to develop is the ability to search large repositories of first-person videos automatically to find a target individual. Much like `ego-surfing' to search on the Internet for our own name, we believe that self-search in first-person videos can empower users to monitor and manage their own personal data. To this end, we develop a video-based self-search technique tailored to first-person videos. Because the appearance of people in first-person videos often comes under motion blur and extreme head/body pose changes (see Fig.~\ref{fig:teaser}), a robust approach beyond what can be accomplished by face recognition alone~\cite{Hjelmas2001,Zhang2010,Zhao2003} is required.

\begin{figure}[t]
\centering
\includegraphics[width=\linewidth]{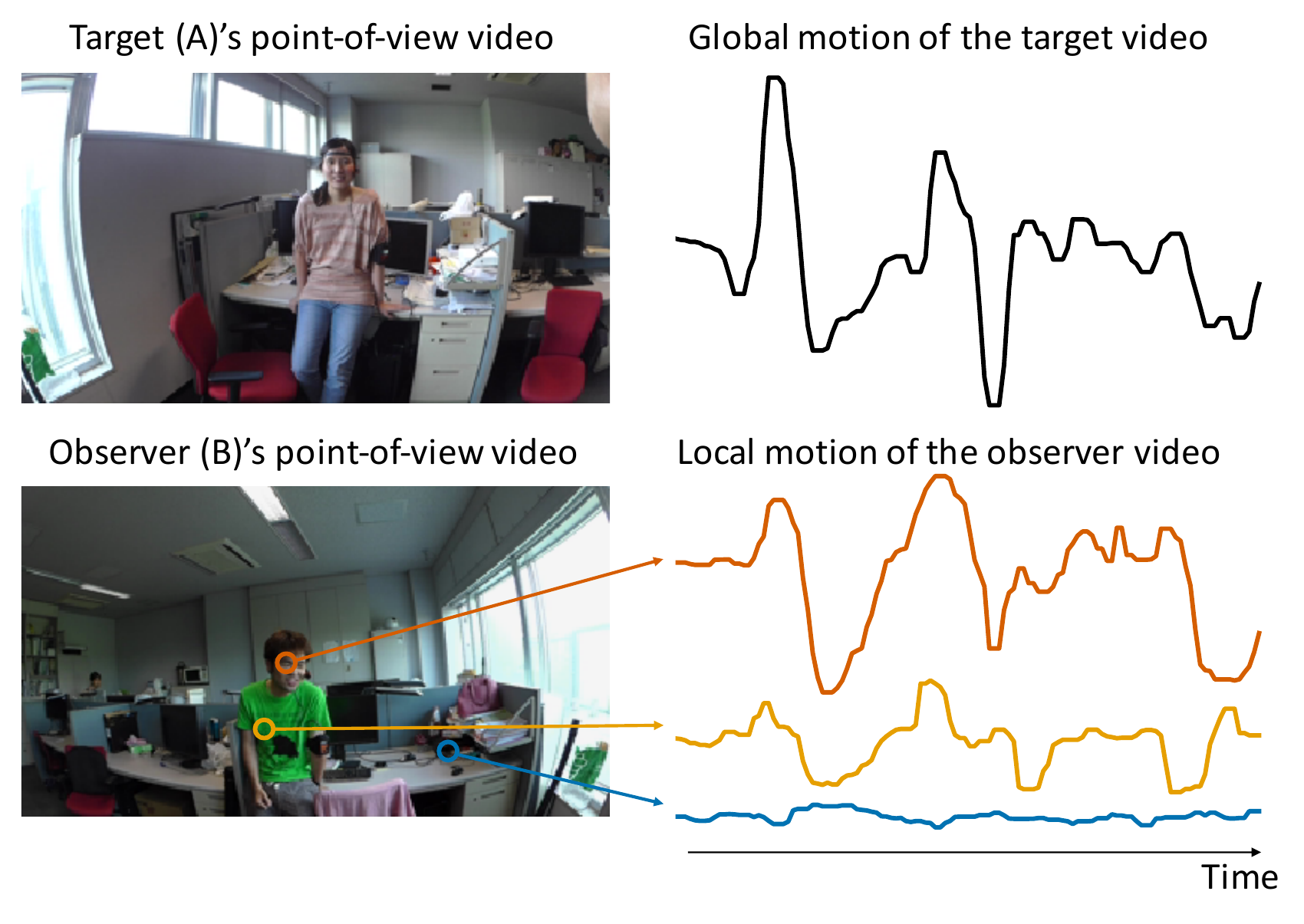}
\caption{Global motion of the target (A)'s points-of-view video and local motions in the observer (B)'s video. The local motion in a target region ({\bf red} line plot) has higher correlation with the global motion ({\bf black} line plot) compared to that in non-target regions ({\bf yellow} and {\bf blue} plots).}
\label{fig:motion}
\end{figure}

To deal with the high variability of self appearance in first-person videos, we propose using motion as our primary feature. The key insight of our work is that a first-person video of a target individual can act as a unique identifier to enable a target-specific search over a repository of first-person videos recorded by others. We give a concrete example in Fig.~\ref{fig:motion}. Consider a case where the target (self) individual A is conversing with another person (whom we will call observer B). When A shakes his head, it induces large global motion (the camera moves according to the black line plot in Fig.~\ref{fig:motion}) in the video. Now, from the perspective of observer B, we expect to see the same shake pattern but in the form of a local motion pattern in B's point-of-view video (we see the target individual A shaking his head at the red circle in Fig.~\ref{fig:motion}). This correlation between the global motion of target A's video and the local motion of observer B's video indicates that these two videos are indeed related. Furthermore, this correlation is expected to increase only in target regions. This illustrates the general insight that the ego-motion of a target is a unique signature that can be used to localize the self in observer videos.

Based on this insight, we develop a novel motion correlation-based approach to search and localize a target individual in a collection of first-person videos. The algorithm takes as input the target's point-of-view video and retrieves as output all instances of that target individual from an observer's point-of-view videos. Our algorithm proceeds as follows. First, densely-sampled point trajectories are generated over all observer videos as target candidates. Second, each trajectory is evaluated to compute their `targetness' based on the correlation between the local motion pattern along the trajectory and the global motion pattern of the target video. Third, a supervised classifier on trajectories refines the targetness scores by taking into account the generic targetness of the trajectories. The targetness evaluation is posed as a binary-class Bayesian inference problem where the likelihood and prior are given by the motion correlation and trajectory classifier, respectively.

Experimental results show that our approach significantly improves the self-search performance over baseline face recognizers. Furthermore, to demonstrate the potential impact that self-search can have on assistive technologies, we applied our method to three tasks: (1) privacy filtering, (2) target video retrieval, and (3) social group clustering.

\subsection*{Related Work}
The idea of searching for a specific person in images or videos has been addressed in several areas of computer vision. One example is \emph{person re-identification} in the context of visual surveillance. An extensive survey of the field can be found in~\cite{Shaogang2014, Vezzani2013, Wang2013}. One common approach for person re-identification is to utilize visual signatures of a specific person, such as color and texture, to find specific individuals in images. Because many approaches presuppose a surveillance scenario, the features and approaches are often adapted for videos captured by a static camera (single point-of-view, constant background, \etc). With the exception of work using active cameras~\cite{Salvagnini2013}, re-identification approaches are not designed to deal with extreme camera motion.

Work in the area of \emph{first-person vision} has utilized person identification as a feature for analyzing human interactions. Many studies have relied on off-the-shelf face detectors and recognizers~\cite{Alletto2014b,Alletto2014,Fathi2012,Lee2012,Poleg2014,Rehg2013,Ye2012,Ye2015,Alletto2015}. In many of these scenarios, the use of face detection is justified since people are engaged in conversation and a wearable camera is relatively stable. More recently, some work has presented a method to identify first-person camera wearers in surveillance videos~\cite{Ardeshir2016,Fan2017}. These methods can work only when target people are detected stably in surveillance videos. In addition, \cite{Ardeshir2016} requires multiple wearable cameras to share their fields-of-view to measure visual similarity between recorded videos. Similarly, scenes around target people have to be observed both in first-person and surveillance videos to enable spatio-temporal analysis in~\cite{Fan2017}. By contrast, our approach does not rely on any person detection and accepts first-person videos that have significantly different fields-of-view with few overlaps.

Another approach for identifying specific individuals is the use of geometric information in 2D~\cite{Hesch2012, Murillo2012} or 3D~\cite{Park2012, Park2013, Park2015}. An accurate geometric map can be used to compute the precise location of individuals with wearable cameras, and the location can be used to estimate visibility in other cameras. However, these approaches often require preliminary scanning of scenes (\eg,~\cite{Park2012}), making it applicable only in pre-recorded and static places.

Finally, there are some attempts to make use of head motion patterns for finding specific persons in videos. Poleg \etal~\cite{Poleg2014} has proposed a person identification method based on the correlation of head motion in first-person videos. Their method however relies on person detection to track the head of each target candidate, making it challenging to perform the identification task reliably when a person is mobile and when significant ego-motion is present. By contrast, our approach does not require person detection but directly examines the correlation over densely-sampled point trajectories. This makes our method perform well even with videos captured under significant head motion. Hoshen and Peleg~\cite{Hoshen2016} have attempted to identify individuals by learning their head motion patterns. In contract to their approach that requires head motion classifiers to be trained \emph{per individual} before identification, our work uses \emph{generic} targetness to guide correlation-based identification, which does not have to be learned for specific individuals.

\subsection*{Contributions}
In this work, we extend our prior work presented in~\cite{Yonetani2015}. To the best of our knowledge,~\cite{Yonetani2015} is the first to address the topic of self-search in first-person videos with significant ego-motion, where face recognition and geometric localization are not applicable (\ie, people with high facial appearance variability, captured by the cameras with significant motion, without any restriction on recorded places). In addition, this work makes the following contributions:
\begin{itemize}
\item Unlike our previous approach~\cite{Yonetani2015}, we introduce a new form of target candidates based on densely-sampled point trajectories in Section~\ref{sec:method}. Compared to the supervoxel-based candidates used in~\cite{Yonetani2015}, trajectory-based candidates are comparably robust against the variability of faces, and they can be generated much faster. Experiments in Section~\ref{sec:experiments} demonstrate the effectiveness of our new approach in terms of both accuracy and efficiency.
\item We propose a two-step search scheme to accelerate targetness evaluation in Section~\ref{sec:upperbound}. An upperbound of correlation-based targetness is rapidly estimated in the first step to limit the number of candidates to apply stable but slow correlation evaluation in the latter step. This makes the overall procedure more efficient while keeping high search performance.
\item In addition to the task of target localization addressed originally in~\cite{Yonetani2015}, we quantitatively evaluated our approach on the tasks of target video retrieval (searching a repository of first-person videos for the videos including target individuals) and social group clustering (finding a group of videos recording the same social interaction) in Section~\ref{sec:experiments}.
\end{itemize}

\section{Proposed Target Search}
\label{sec:method}
Assume that we are given a repository of first-person videos that captured people's interactions in various places. Our goal is to search for a target person (\ie, the self) across the repository. More specifically, we wish to evaluate how likely each video includes the target individual and to localize all the instances of the target in the video. In what follows, we consider two types of videos: videos recorded (1) by the target (\emph{target video}) and (2) by observers (\emph{observer videos}). 

\subsection{Overall Pipeline}
\begin{figure*}[t]
\centering
\includegraphics[width=\linewidth]{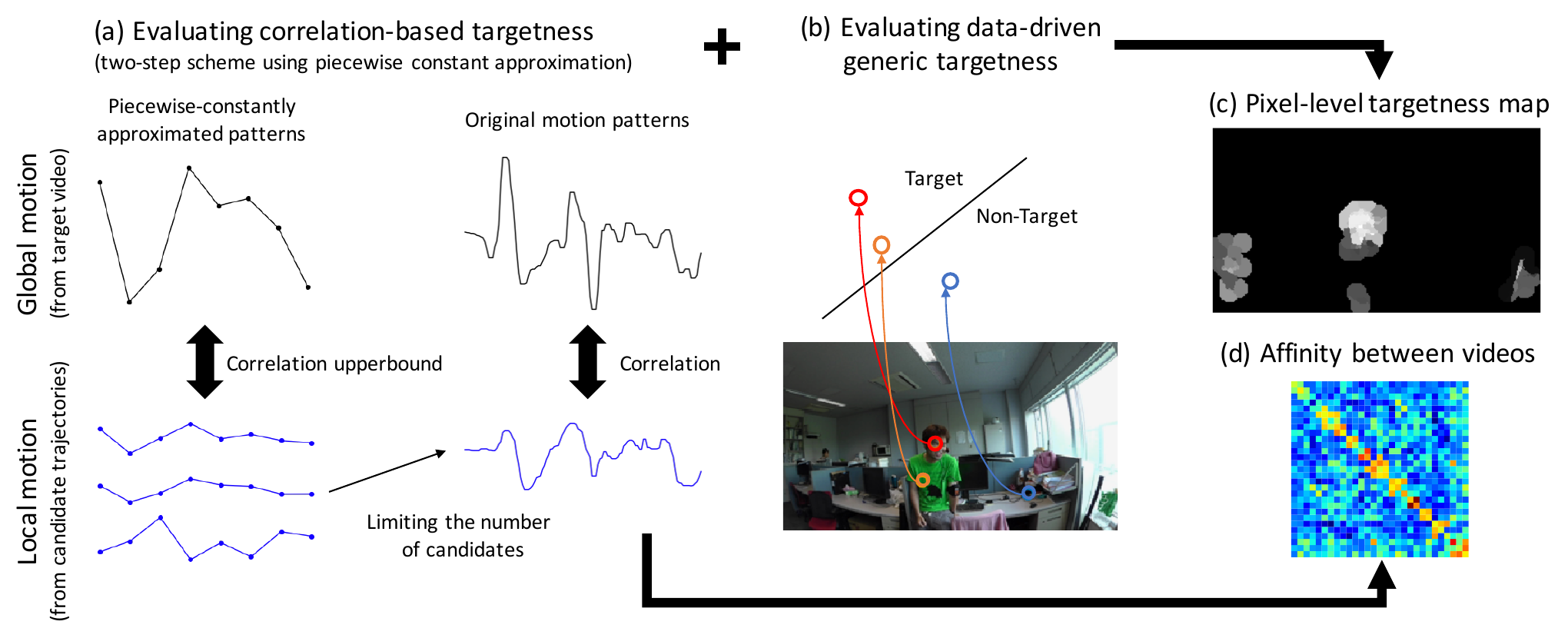}
\caption{Overview of the proposed target search. Given target candidates extracted from observer videos, we evaluate their (a) correlation-based targetness (Section~\ref{subsubsec:correlation_targetness}) and (b) data-driven generic targetness (Section~\ref{subsubsec:generic_targetness}) and combine them to construct (c) a pixel-level targetness map (Section~\ref{subsec:perpixel}). Correlation-based targetness is evaluated efficiently with the two-step scheme presented in Section~\ref{sec:upperbound} and also used to measure (d) the affinity between target and observer videos as shown in Section~\ref{subsec:affinity}.}
\label{fig:overview}
\end{figure*}

This section first describes how the proposed method finds specific target individuals in observer videos. As we will present in Section~\ref{subsec:generating_trajectories}, target candidates are first generated from observer videos. Overall, we evaluate \emph{targetness} of these candidates, that is how likely candidates are to be a part of target individuals (Section~\ref{subsec:evaluating_targetness}). As illustrated in Figure~\ref{fig:overview}(a), local motion of these candidates are compared against the global motion of a target video to measure their correlation (Section~\ref{subsubsec:correlation_targetness}). We refer to this correlation score as \emph{correlation-based} targetness, which will be evaluated with an efficient two-step scheme (Section~\ref{sec:upperbound}). At the same time, we compute \emph{data-driven generic} targetness of candidates using a classifier that learns various traits of candidates to be a target (Figure~\ref{fig:overview}(b) and Section~\ref{subsubsec:generic_targetness}).

In Section~\ref{subsec:perpixel}, we combine correlation-based targetness and data-driven generic targetness to provide a pixel-level targetness map shown in Figure~\ref{fig:overview}(c). We also make use of the correlation-based targetness to measure the affinity between target and observer videos (Figure~\ref{fig:overview}(d)) in Section~\ref{subsec:affinity}, which indicates how likely two camera wearers interact with each othere

\subsubsection{Generating target candidates}
\label{subsec:generating_trajectories}
Generating good target candidates is essential to achieve robust and efficient search. Poleg \etal~\cite{Poleg2014} extract target candidates via person detection. However, person detection often becomes difficult when heavy occlusion and motion blur are present on faces or when the faces are extremely small in observer videos. Our prior work~\cite{Yonetani2015} instead used a supervoxel hierarchy~\cite{Xu2012a, Xu2012b}. While this approach has enabled robust search against the variability of faces, hierarchical supervoxel segmentation requires rather long computation time making it difficult to apply on a large-scale repository of first-person videos.

As a more efficient approach, we present a new target candidate based on densely-sampled point trajectories. That is, feature points detected in a certain frame are tracked over time to serve as the candidates. Target candidates generated in this way allow us to avoid evaluation in obvious background regions (\eg, texture-less walls and floors), making the overall search procedure more efficient than the supervoxel-based approach used in~\cite{Yonetani2015}. They are also comparably robust against the facial appearance variability because they rely only on the motion estimation.

Following~\cite{Wang2013}, feature points are sampled with a uniform step $e_W$ and selected with a threshold based on the good-feature-to-track criterion~\cite{Shi1994}. Dense optical flow fields such as~\cite{Farneback2003} are then estimated to track the points. While~\cite{Wang2013} limits the length of trajectories to ensure that is short enough to avoid drifting, we want trajectories that are as long as possible to compute a motion correlation stably, as suggested in~\cite{Poleg2014}. Therefore, we instead set the minimum and maximum lengths to the trajectories, $[L\sub{\min}, L\sub{\max}]$, and continue tracking as long as the trajectories are shorter than $L\sub{\max}$ and the tail of each trajectory satisfies the good-feature-to-track criterion. Trajectories shorter than $L\sub{\min}$ are just unused for target candidates.

\subsubsection{Evaluating targetness of candidates}
\label{subsec:evaluating_targetness}
Given densely-sampled trajectories as target candidates, we then evaluate their targetness using target video as a search query. In what follows, we introduce a mathematical formulation of the targetness evaluation.

Let $\mathcal{X}=\{X^{(1)},\dots,X^{(N)}\}$ be a set of candidate trajectories. Each trajectory $X^{(i)}$ is defined by a sequence of points $X^{(i)}=(\vct{x}^{(i)}_{1}, \dots, \vct{x}^{(i)}_{l^{(i)}})$, where $\vct{x}_t \in \mathbb{R}^{2}_+$ is a pixel at the $t$-th frame, starting at the $b^{(i)}$-th frame and lasting for $l^{(i)}$ frames. For each candidate $X^{(i)}$, we introduce a binary assignment variable $a_{X^{(i)}} \in \{1, 0\}$ that takes $a_{X^{(i)}}=1$ if $X^{(i)}$ is in a target region. Then, the targetness of $X^{(i)}$ given a target video $\mathcal{V}\sub{G}$ is defined by a posterior probability $P(a_{X^{(i)}}\mid \mathcal{V}\sub{G})$. We further decompose this posterior using Bayes' rule to obtain the following Eq.~(\ref{eq:bayes}).
\begin{equation}
P(a_{X^{(i)}} \mid \mathcal{V}\sub{G}) \propto P(\mathcal{V}\sub{G}\mid a_{X^{(i)}})P(a_{X^{(i)}}).
\label{eq:bayes}
\end{equation}
As will be explained shortly, we will give the likelihood $P(\mathcal{V}\sub{G}\mid a_{X^{(i)}})$ by correlation-based targetness and the prior term $P(a_{X^{(i)}})$ by data-driven generic targetness.


\subsection{Correlation-based Targetness}
\label{subsubsec:correlation_targetness}
Given a target video as a search query, each target candidate obtains correlation-based targeteness: how much its local motion is correlated with the referential global motion of the target video. We follow~\cite{Poleg2014} and evaluate the zero-mean normalized cross correlation (ZNCC) independently along horizontal and vertical directions to average them.

Motion estimation for target and observer videos proceeds as follows. In the target video, we first compute sparse optical flows by~\cite{Lucas1981}. We follow~\cite{Wang2013} and assume that two consecutive frames are related by a homography. Then, the homography estimated from the flows is used to compute a global motion vector for every pixel. Finally, we average the global motion vectors for each frame to obtain a global motion pattern as a sequence of two dimensional (\ie, horizontal and vertical) vectors. As for the observer videos, dense optical flow fields are estimated by~\cite{Farneback2003} in addition to the per-pixel global motion vectors. A local motion vector for each pixel is then computed by subtracting the global motion vector from the optical flow. Finally, a local motion pattern on a candidate trajectory is obtained by referring to the local motion at each point of the trajectory. To cope with motion estimation error, we apply a median filter on global and local motion patterns as post-processing.

Now, denote the local motion pattern of a candidate trajectory $X^{(i)}$ as a sequence of two dimensional vectors, $\vct{m}^{(i)} = \left((u^{(i)}_1, v^{(i)}_1), \dots, (u^{(i)}_{l^{(i)}}, v^{(i)}_{l^{(i)}})\right)$, where $u_t^{(i)}$ and $v_t^{(i)}$ are horizontal and vertical elements of motion, respectively. Likewise, we consider the global motion pattern $\vct{M}=\left((U_1, V_1), \dots, (U_{l^{(i)}}, V_{l^{(i)}})\right)$ adaptively cropped according to the interval $[b^{(i)}, b^{(i)} + l^{(i)}]$ ($U_t$ and $V_t$ are also horizontal and vertical elements, respectively). Note that horizontal and vertical elements of $\vct{m}^{(i)}$ and $\vct{M}$ are independently normalized to have zero mean and unit variance beforehand, and the vertical elements of global motion, $V_t$, are inverted as they are inverted to $v^{(i)}_t$ (e.g., head motion by nodding down appear as upper global motion in a target video). Then, the correlation between $\vct{m}^{(i)}$ and $\vct{M}$ is defined by averaging horizontal and vertical ZNCCs as follows:
\begin{equation}
C(\vct{m}^{(i)}, \vct{M}) = \frac{1}{2}\left(\frac{\sum_t u^{(i)}_t U_t}{l^{(i)}} + \frac{\sum_t v^{(i)}_t V_t}{l^{(i)}}\right).
\label{eq:corr}
\end{equation}
Finally, we scale $C(\vct{m}^{(i)}, \vct{M})$ into the range of $[0, 1]$ to obtain the correlation-based targetness. Following~\cite{Yonetani2015}, we use a sigmoid function for scaling; namely, $P(\mathcal{V}\sub{G}\mid a_{X^{(i)}}) \triangleq (1 + \exp(-C(\vct{m}^{(i)}, \vct{M})))^{-1}$.

\subsection{Efficient Search Using Correlation Upperbounds}
\label{sec:upperbound}
We consider accelerating the evaluation of correlation-based targetness as one of our main extensions over~\cite{Yonetani2015}. This acceleration is particularly critical because targetness evaluation needs to be performed against all videos in a repository as many times as a new query (\ie, target) video is made. In this section, we introduce an efficient evaluation scheme of correlation-based targetness by estimating its upperbound.

\subsubsection{Two-step evaluation of targetness}
As can be seen in Eq.~(\ref{eq:corr}), the time complexity of evaluating correlation-based targetness is linearly proportional to the temporal length of target candidates (\ie, $\mathcal{O}(l^{(i)})$ for each candidate). Namely, the longer time we need to track candidates to stably calculate a correlation, the longer time we need to evaluate their targetness.

To solve this problem, our proposal is to limit the number of candidates to perform a stable but slow correlation evaluation. More specifically, we introduce the piecewise constant approximation of motion patterns to estimate an upperbound of correlation scores between the patterns. This upperbound estimation for each candidate can be done in constant time regardless of the candidate length, making it possible to avoid immediately many unnecessarily evaluations of candidates resulting in a low correlation score.

Our scheme is summarized in the following two steps (see also Figure~\ref{fig:overview} (a)): (1) approximating motion patterns into $K$ constant pieces to estimate an upperbound of correlation scores and (2) choosing the candidates having the top $P$-percentile high upperbound score to compute their actual correlations. Now, the time complexity improves to $\mathcal{O}(K)$ for each candidate and $\mathcal{O}(l^{(i)})$ for the only $P$ percentage of all the candidates (usually $K\ll l^{(i)}$). Note that the scheme is inspired by the GEMINI (generic multimedia indexing) framework~\cite{Faloutsos1994,Keogh2001} commonly used in the field of time-series data mining.

\subsubsection{Estimating an upperbound of correlation}
We first define piecewise constant approximation. Let $\vct{u}=(u_1,\dots,u_l)$ and $\vct{U}=(U_1,\dots,U_l)$ be horizontal patterns of local and global motion, respectively (we omit the index of local motion patterns without loss of generality). We assume that $\vct{u}$ and $\vct{U}$ are normalized to have zero mean and unit variance. $K$-piecewise constant approximation of the local motion pattern $\vct{u}$ results in a $K$-dimensional pattern, $\bar{\vct{u}}=(\bar{u}_1,\dots,\bar{u}_K)$, where $\bar{u}_k$ is defined as follows:
\begin{equation}
\bar{u}_k = \frac{K}{l}\sum_{t=(k-1)\frac{l}{K}+1}^{k\frac{l}{K}}u_t.
\end{equation}
Note that $K$ is set to be a divisor of $l$ and is constant regardless of the length of original patterns\footnote{In practice, we trim the last several frames of each trajectory so that $l$ is a multiple of $K$.}. In the same way, we can also obtain the approximation for the global motion pattern $\vct{U}$ as $\bar{\vct{U}}=(\bar{U}_1,\dots,\bar{U}_K)$.

As proved in~\cite{Keogh2001}, the Euclidean distance between two piecewise constants provides a lower-bound of their actual Euclidean distance. Namely, the following inequality holds: 
\begin{equation}
\frac{1}{l}\sum_t (u_t - U_t)^2 \geq \frac{1}{K}\sum_k(\bar{u}_k - \bar{U}_K)^2,
\label{eq:lb}
\end{equation}
where the equality holds for $K=l$. Therefore, an upperbound of ZNCC between $\vct{u}$ and $\vct{U}$ can be derived as follows:
\begin{equation}
\frac{1}{l}\sum_t u_t U_T \leq \frac{1}{K}\sum_k \bar{u}_k \bar{U}_k + \left(1-\frac{\sigma^2_{\bar{\vct{u}}} + \sigma^2_{\bar{\vct{U}}}}{2}\right),
\label{eq:ub}
\end{equation}
where $\sigma^2_{\bar{\vct{u}}}$ and $\sigma^2_{\bar{\vct{U}}}$ are the variance of piecewise constants. By letting $\bar{\vct{v}}$ and $\bar{\vct{V}}$ be the piecewise constant approximation of vertical patterns of local and global motion, and by letting $\sigma^2_{\bar{\vct{v}}}$ and $\sigma^2_{\bar{\vct{V}}}$ be their variances, the upperbound of Eq.~(\ref{eq:corr}) can be derived as follows:
\begin{eqnarray}
C(m^{(i)}, M) &\leq& \frac{1}{2}\left(\frac{\sum_k \bar{u}^{(i)}_k \bar{U}_k}{K} + \frac{\sum_k \bar{v}^{(i)}_k \bar{V}_k}{K}\right) + Z. \\
Z&=&1-\frac{\sigma^2_{\bar{\vct{u}}} + \sigma^2_{\bar{\vct{U}}} + \sigma^2_{\bar{\vct{v}}} + \sigma^2_{\bar{\vct{V}}}}{4}.
\label{eq:corr_ub}
\end{eqnarray}
In this upperbound, $Z$ increases up to $1$ as $l$ becomes larger than $K$. This property works well because the top $P$-percentile target candidates preferentially include the ones seen by observers for a longer time, and they are more likely to be actual target instances in practice. In the following experiments, we demonstrate that the proposed two-step scheme successfully reduces computation time while providing comparable localization performance to where the actual correlation is computed for all the candidates.

\subsubsection*{Deriving an upperbound of correlation}
Here we introduce how to derive Eq.~(\ref{eq:ub}) from Eq.~(\ref{eq:lb}). By expanding the LHS of Eq.~(\ref{eq:ub}), we obtain
\begin{eqnarray}
\frac{1}{l}\sum_t (u_t - U_t)^2 &=&
\frac{1}{l}\sum_t(u_t^2 - 2 u_tU_t + U_t^2 )\\
&=&2(1 - \frac{1}{l}\sum_tu_t U_t).
\label{eq:expand1}
\end{eqnarray}
In Eq.~(\ref{eq:expand1}), recall that $\vct{u}$ and $\vct{U}$ are normalized to have zero mean and unit variance, and thus $\frac{1}{l}\sum_t u_t^2=\frac{1}{l}\sum_t (u_t^2 - \mbox{mean}(\vct{u})) = \sigma^2_{\vct{u}} = 1$, and $\frac{1}{l}\sum_tU_t^2 = 1$. Likewise, we can expand the RHS of Eq.~(\ref{eq:lb}) as follows:
\begin{equation}
\frac{1}{K}\sum_k (\bar{u}_k - \bar{U}_k)^2 = \sigma^2_{\bar{\vct{u}}} + \sigma^2_{\bar{\vct{U}}} - \frac{2}{K}\sum_k\bar{u}_k \bar{U}_k.
\label{eq:expand2}
\end{equation}
By substituting Eq.~(\ref{eq:expand1}) and Eq.~(\ref{eq:expand2}) into Eq.~(\ref{eq:lb}) we get Eq.~(\ref{eq:ub}).

\subsection{Data-driven Generic Targetness}
\label{subsubsec:generic_targetness}
The prior $P(a_{X^{(i)}})$ in Eq.~(\ref{eq:bayes}) is introduced to consider generic targetness of a candidate trajectory $X^{(i)}$. For example, trajectories that track a skin-color point are more likely to be a part of facial regions, and stationary trajectories often come from a background region. This prior aims to learn such traits from many pairs of observer videos and corresponding masks annotating target regions. While manual annotations of target persons are required here, this prior is independent of specific individuals and backgrounds. Therefore, the learning of the prior needs to be carried out only once and is not necessary for each target.

Specifically, we first extract positive and negative feature samples from target and non-target regions, respectively, and then train a binary classifier. This classifier produces a posterior probability of target candidates belonging to the target class, \ie, $P(a_{X^{(i)}}=1)$. To capture a trait of generic targetness, we extract color (in the HSV space) and local motion for each pixel on candidate trajectories. The mean and standard deviation of the color and motion are calculated to serve as features. We also use the temporal length of trajectories as a feature.

\subsection{Pixel-level Targetness Map}
\label{subsec:perpixel}
$P(a_{X^{(i)}} \mid \mathcal{V}\sub{G})$ in Eq.~(\ref{eq:bayes}) describes how likely each trajectory is to be a part of target instances. In order to localize target regions in observer videos as we do in Figure~\ref{fig:teaser}, we need to extend this trajectory-level evaluation to be a pixel-level targetness map shown in Figure~\ref{fig:overview}(c). Unlike supervoxel candidates used in~\cite{Yonetani2015}, trajectory candidates do not cover an entire video frame. This requires some interpolation techniques to generate the pixel-level map. We here introduce a simple nearest-neighbor approach to this problem. Intuitively, a certain pixel will take the targetness of its nearest neighbor candidate trajectory if the candidate is sufficiently close to the pixel.

Like a binary assignment $a_{X^{(i)}}$ defined for each candidate $X^{(i)}$, we consider a binary assignment variable $a_{\vct{x}_t}$ for each pixel $\vct{x}_t$. That is, $a_{\vct{x}_t}=1$ if $\vct{x}_t$ corresponds to target regions in an observer video. The targetness for pixel $\vct{x}_t$ is defined by the posterior probability of $a_{\vct{x}_t}=1$ given a target video $\mathcal{V}\sub{G}$. The nearest-neighbor rule to interpolate $a_{\vct{x}_t}$ from $a_{X^{(i)}}$ is defined as follows:
\begin{equation}
P(a_{\vct{x}_t} | \mathcal{V}\sub{G}) \triangleq
\begin{cases}
P(a_{X^{(\hat{i})}} \mid \mathcal{V}\sub{G}) & \mbox{if}\; D(\vct{x}_t, X^{(\hat{i})}) \leq r,  \\
0 & \mbox{otherwise,}
\label{eq:posterior}
\end{cases}
\end{equation}
where $\hat{i} = \arg\min_i D(\vct{x}_t, X^{(i)})$ is given by an edge-guided geodesic distance. Specifically, we first construct a graph where each pixel is referred to as a node that connects to its adjacent eight pixels with an edge. Each edge is given its weight by a difference in the two pixel values. Then, $D(\vct{x}_t, X^{(i)})$ is defined by a cost of the shortest path between $\vct{x}_t$ and $\vct{x}^{(i)}_{t - b^{(i)} + 1}$ if $b^{(i)}\leq t < b^{(i)} + l^{(i)}$, and $\infty$ otherwise. The radius threshold $r$ is defined to be twice as large as the trajectory sampling step $e_W$ to cover pixels sufficiently between candidates. This way, we expect pixel-wise target maps to highlight target regions surrounded by strong edges. We will use this per-pixel targetness map for the task of target localization in Section~\ref{subsec:localization_experiment}.

\subsection{Affinity between First-Person Videos}
\label{subsec:affinity}
In addition to the localization of target regions in particular observer videos, another important ability of a target search is to retrieve videos that include a target individuals from a collection of first-person videos. As we assume in this work that first-person videos are recorded during interactions, such videos are highly likely to be taken by the people involved in the same interaction group. To take this into account, we propose an affinity between two first-person videos that describes how likely two camera wearers interact with each other for target video retrieval.

First, we define how likely a person $p$ recording the video $\mathcal{V}_p$ is to interact with a person $q$ recording the video $\mathcal{V}_q$ by $A(\mathcal{V}_q\mid \mathcal{V}_p)$. We expect that, the longer $p$ appears in the video $\mathcal{V}_q$, the more likely $p$ is interacting with $q$. Given a set of candidates $\mathcal{X}=\{X^{(1)},\dots,X^{(N)}\}$ generated from $\mathcal{V}_q$, $A(\mathcal{V}_q\mid \mathcal{V}_p)$ is defined as follows: 
\begin{equation}
A(\mathcal{V}_q\mid \mathcal{V}_p) = \max_{i}P(\mathcal{V}_p\mid a_{X^{(i)}})\cdot S(l^{(i)}; \mu_l),
\label{eq:aff1}
\end{equation}
where $S(l^{(i)}; \mu_l)=\left(1 + \exp(-l^{(i)} + \mu_l)\right)^{-1}$ is a sigmoid function with the center at $\mu_l$ to weigh each candidate based on its length. We also point out that correlation-based targetness $P(\mathcal{V}_p\mid a_{X^{(i)}})$ is more necessary to compute the affinity than data-driven generic targetness $P(a_{X^{(i)}})$ because $P(a_{X^{(i)}})$ encourages \emph{any} candidates equally regardless of $\mathcal{V}_p$, making affinities more similar to each other. 

Another important observation is that, when a target person is interacting with an observer, the observer can also be present in the first-person video recorded by the target. Namely, at least one of two people is expected to be seen in the other video if they are interacting with each other. Therefore, we define the affinity between two videos, $A(\mathcal{V}_p, \mathcal{V}_q)$, as follows:
\begin{equation}
A(\mathcal{V}_p, \mathcal{V}_q) = \max\left(A(\mathcal{V}_p\mid \mathcal{V}_q) , A(\mathcal{V}_q\mid \mathcal{V}_p)\right).
\label{eq:aff2}
\end{equation}

This symmetric affinity can indeed work better than the asymmetric targetness defined in Eq.~(\ref{eq:aff1}), as will be demonstrated in our experiments. We also show in the experiments that this affinity can be further applied to find groups of videos recording the same social interaction from a collection of first-person videos.
\section{Experiments}
\label{sec:experiments}
In this section, we evaluate the proposed approach on our new dataset introduced in \cite{Yonetani2015}\footnote{Our dataset and codes will be available on the project page http://yonetaniryo.github.io/corrsearch/.} and CMU-group first-person video dataset (CMU dataset) used in \cite{Arev2014, Park2012, Park2013}. 

\subsection{Datasets}
\begin{figure*}[t]
\centering
\includegraphics[width=.9\linewidth]{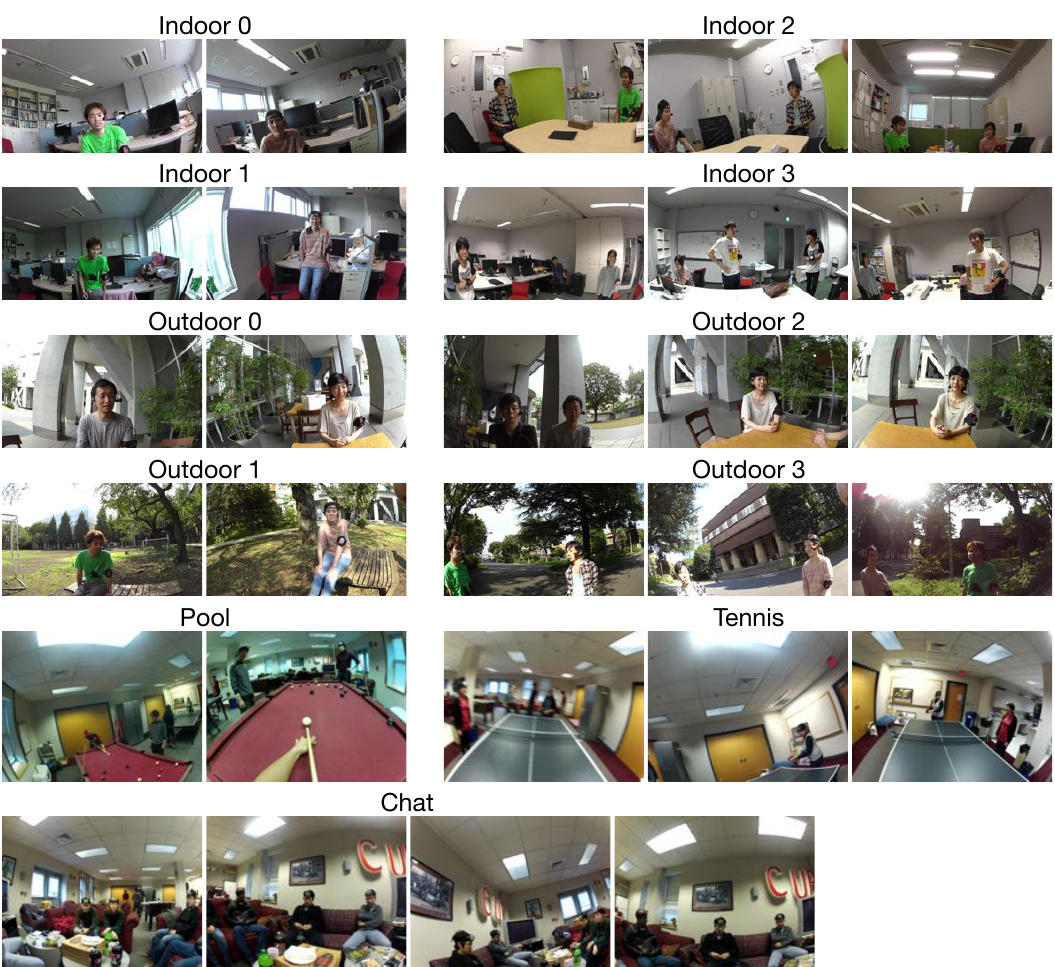}
\caption{Representative frames of interaction scenes in our dataset and CMU dataset.}
\label{fig:example}
\end{figure*}

Our dataset was collected in eight different interaction scenes. The number of participants equipped with a wearable camera was two or three for each scene. These participants stayed at the same position but often changed their poses. Interactions were recorded at 60 fps, 30 sec in four indoor ({\bf Indoor 0} -- {\bf Indoor 3}) and four outdoor scenes ({\bf Outdoor 0} -- {\bf Outdoor 3}). On CMU dataset, 11 participants formed groups to play pool, to play table tennis, to sit on couches to chat, or to talk with each other at a table. They changed their poses and positions frequently and often disappeared from observer videos, standing for a more challenging scenario than our dataset. For the nine videos available for analysis, we used 3861st -- 5300th frames (30 sec at 48 fps), which comprised {\bf Pool}: two people played pool, {\bf Tennis}: three played table tennis, and {\bf Chat}: the remaining four chatted on a couch. In total, 29 videos comprising 11 different scenes were used (see also Figure~\ref{fig:example}).

We manually annotated image regions corresponding to a target person's head every 0.5 second. This is because the local motion corresponding to ego-motion should be observed in the region of the head. These annotations also served as a supervised label to learn the prior $P(a_{X^{(i)}})$. We used the linear discriminant analysis by following \cite{Yonetani2015} because it performed the best but any other classifier could work as well. Because we evaluated two completely different datasets, we used one for training the prior to test the other. It ensured that people and scenes in test subsets did not appear in training ones. 

\subsection{Implementation Details}
\label{sec:implementation}
In our experiments, we resized video frames into 320x180 to complete an overall procedure in a reasonable time. The spatial step $e_W$ to sample trajectories was set to four pixels. The good-feature-to-track criterion, \ie, the minimum eigenvalue of the auto-correlation matrix, was set to $10^{-4}$. Every four frames, new trajectories were sampled at the locations that satisfy the aforementioned criteria and is not covered by current trajectories. Smaller values for $e_W$ and the good-feature-to-track criterion could allow us to use more densely-sampled trajectories while we found that such settings slightly improved performance at a cost of longer computation time. $L\sub{min}$ and $L\sub{max}$ were set to 64 and 1024 frames, respectively. The average trajectory length $\mu_l$ in Sec.~\ref{subsec:affinity} was set to 220.14 frames, larger than the 200 frames that \cite{Poleg2014} claimed to ensure low false-positive rates.

\subsection{Evaluation on Target Localization}
\label{subsec:localization_experiment}
We first adopted our approach to a target localization task. Namely, given a pair of target and observer videos (in what follows, we refer to this pair as \emph{session}), we evaluated how much per-pixel targetness maps in Eq.~(\ref{eq:posterior}) were correlated with ground truth masks. Among the 52 sessions in total, we did not use four sessions (two in {\bf Outdoor 2} and another two in {\bf Chat}), where the target and observer sat down side by side and hardly looked at each other.

To show how the proposed correlation-based and data-driven generic targetness work, we evaluated performances on the following four methods.
\begin{itemize}
\item {\bf Proposed}: the proposed method with the combination of correlation-based and data-driven generic targetness without the two-step targetness evaluation.
\item {\bf Corr-only}: a degraded version of the proposed method using the only correlation-based targetness (\ie, $P(a_{X^{(i)}}=1) = 1$ for all candidates).
\item {\bf Generic-only}: another degraded method that uses the only data-driven generic targetness and sets $P(\mathcal{V}\sub{G}\mid a_{X^{(i)}}=1) = 1$ for all candidates.
\item {{\bf YKS}}: the method proposed in our prior work~\cite{Yonetani2015}. Instead of densely-sampled point trajectories, supervoxel hierarchies were used for target candidates.
\end{itemize}
We also compared the following two methods to show the effectiveness of our efficient search presented in Section~\ref{sec:upperbound}. 
\begin{itemize}
\item {\bf Proposed-ES}: the proposed efficient search using an upperbound of correlation scores. The top 25 percent candidates were chosen to evaluate actual correlations.
\item {\bf ES (Generic)}: baseline efficient approach that picks the top 25 percent of trajectories in terms of the generic targetness score $P(a_{X^{(i)}} = 1)$ instead of the correlation upperbounds.
\end{itemize}
For these methods, we fixed hyper-parameters to be $P=25$ (\ie, the percentage of candidates to compute actual correlation scores) and $K=32$ (\ie, the number of pieces used in the piece-wise constant approximation) and later discussed how performances changed for different choices of these parameters.

In addition, several off-the-shelf face detectors and recognizers served as a baseline. We used the mixtures-of-tree based facial landmark detector~\cite{Zhu2012} ({\bf ZR}) and the Haar-cascade face detector~\cite{Viola2001} ({\bf VJ}). {\bf VJ} combined frontal and profile face models to permit head pose variations. We also utilized face recognizers based on local binary pattern histograms~\cite{Ahonen2006} ({\bf LBPH}) and the Fisher face~\cite{Belhumeur1997} ({\bf FisherFace}). These two recognizers learned target faces from different sessions (for instance, to learn a specific person shown in Indoor 0, we used that person's face shown in Indoor 1, 2, 3 and Outdoor 0, 1, 2, 3) and ran on the detection results of {\bf VJ}. Note that these face detectors and recognizers required full resolution videos (1920x1080 for out dataset and 1280x960 for the CMU dataset) while our algorithm was able to run on videos of size 320x180 or 320x240.

We compared these methods based on the area under the receiver-operator characteristic curve (AUC) score and the average precision (AP) score based on the pixel-wise comparisons between targetness maps and ground-truth annotations. Note that we computed AP scores of targetness-based methods (\ie, {\bf Proposed}, {\bf Corr-only}, {\bf Generic-only}, {\bf YKS}, {\bf Proposed-ES}, and {\bf ES (Generic)}) after binarizing targetness maps with their median value. This post-processing was necessary to enable a fair comparison against AP scores of face detectors and recognizers that provided binary outputs.

\subsubsection*{Results}
\begin{figure*}[t]
\centering
\includegraphics[width=\linewidth]{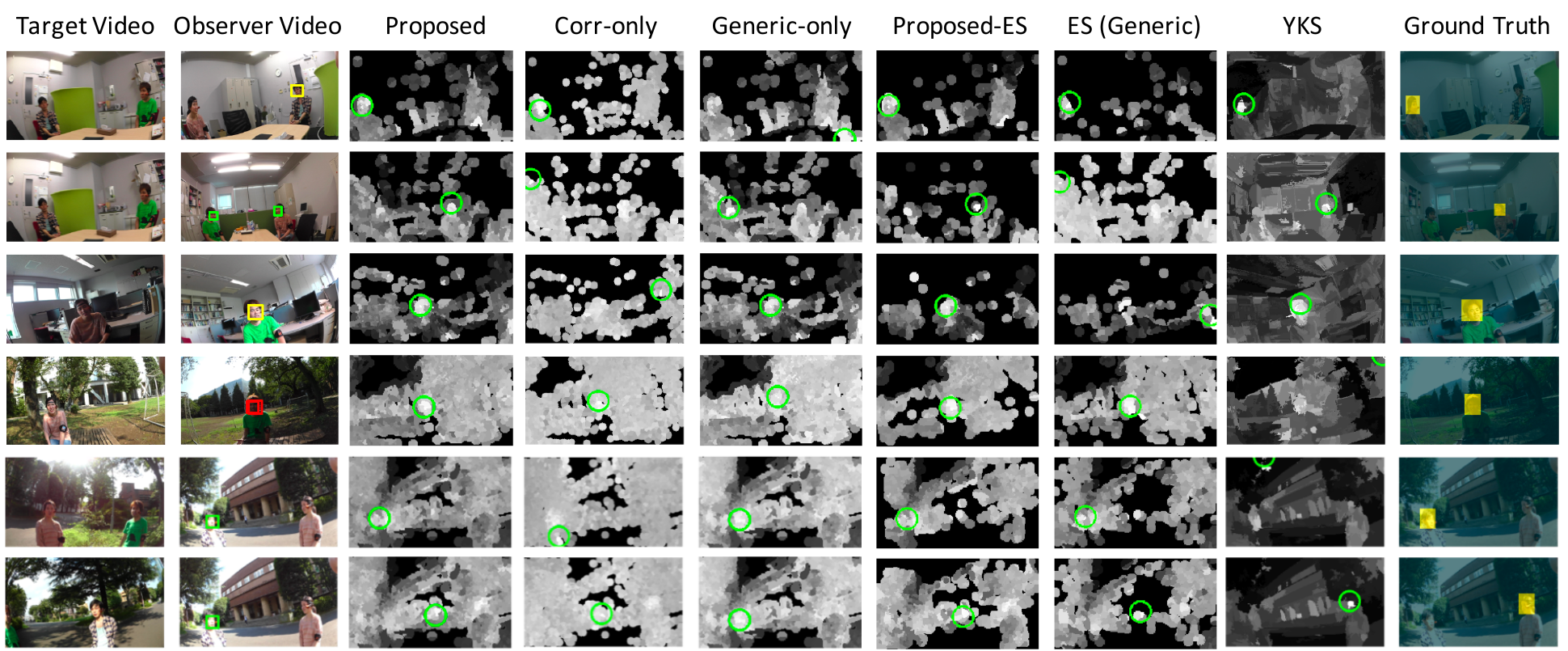}
\caption{Target localization results on our dataset. The target candidate with the highest targetness score is specified by the green circle. Results for {\bf VJ}, {\bf LBPH}, {\bf FisherFace} were respectively shown in green, red, and yellow rectangles in the second column.}
\label{fig:result_yks}
\end{figure*}

\begin{figure*}[t]
\centering
\includegraphics[width=\linewidth]{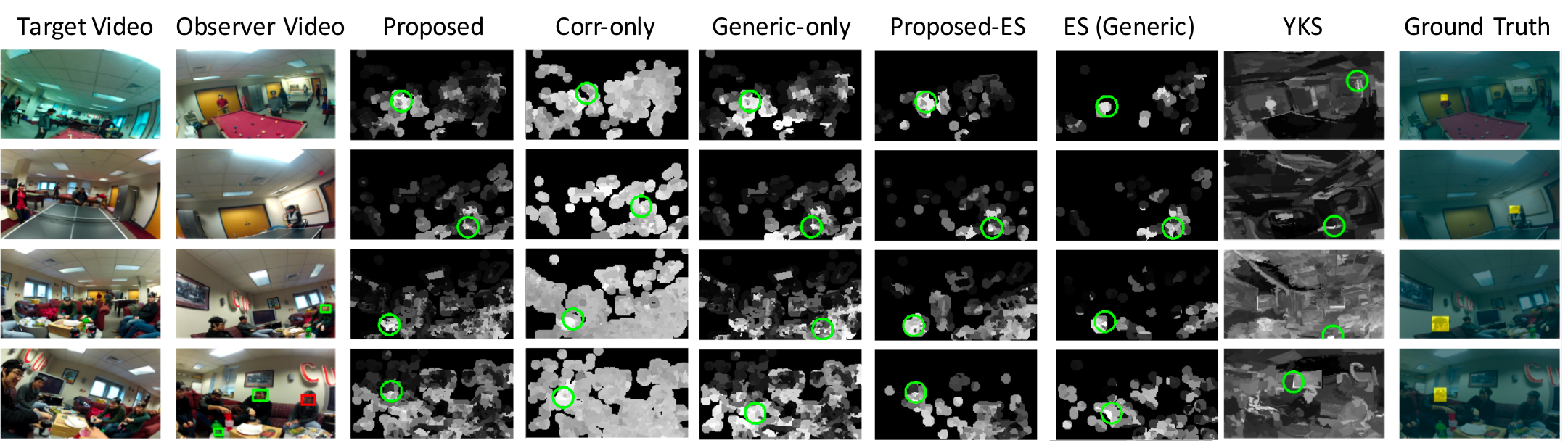}
\caption{Target localization results on the CMU dataset. The target candidate with the highest targetness score is specified by the green circle. Results for {\bf VJ}, {\bf LBPH}, {\bf FisherFace} were respectively shown in green, red, and yellow rectangles in the second column.}
\label{fig:result_cmu}
\end{figure*}

Figure~\ref{fig:result_yks} presents some target localization results on our dataset. In the first and second rows of Figure~\ref{fig:result_yks}, {\bf Proposed} successfully located a target person (the woman in the pink shirt) in two different points-of-view (green circles in the figure). In the third and fourth rows, the target person (the man in the green shirt) was found regardless of the lighting conditions, while none of the two face recognizers could recognize both cases. The bottom two rows show challenging cases where all the methods were not able to distinguish two people in the same observer video (\ie, {\bf Proposed} and {\bf Proposed-ES} detected a correct person on the fifth case while {\bf ES (Generic)} and {\bf YKS} worked only on the sixth case). The comparison between {\bf Proposed}, {\bf Corr-only}, and {\bf Generic-only} indicates that the correlation-based and generic targetness work complementary and are combined effectively in the proposed method. Figure~\ref{fig:result_cmu} shows results on the CMU dataset. Target candidates in the background regions sometimes obtained the highest targetness score incorrectly using {\bf YKS}~\cite{Yonetani2015}. Because our trajectory-based approaches did not evaluate obvious background regions, {\bf Proposed} worked more stably in such cases.

Table~\ref{tab:auc_yks} and Table~\ref{tab:auc_cmu} describe AUC and AP scores. On average, {\bf Proposed} and {\bf Proposed-ES} outperformed the other baseline methods in terms of AUC scores. These proposed methods also showed better AP scores when scenes involve three people (see ``Average (3 people)'' in Table~\ref{tab:auc_yks}) and when people frequently changed their poses and positions (CMU dataset shown in Table~\ref{tab:auc_cmu}). Face detectors~\cite{Viola2001,Zhu2012} worked better only when people are stably observed in videos (\eg, the sessions with two people where each frame contained no other individuals than the target one, as shown in ``Average (2 people)'' in Table~\ref{tab:auc_yks}) and completely failed to detect people in the CMU dataset. Face recognizers~\cite{Ahonen2006,Belhumeur1997} eliminated some incorrect detections of faces but failed to recognize target faces. One reason is the high variability of facial appearances despite the limited number of training face samples for each individual. On the other hand, the proposed method does not require any person-specific training and thus works well under such conditions.

\begin{table*}[t]
\centering
\caption{AUC and AP scores averaged over each scene in our dataset. See Section~\ref{subsec:localization_experiment} for more details on each method.}
\label{tab:auc_yks}
\scalebox{0.85}{
\begin{tabular}{cccccccccccc}
\toprule
\multirow{2}{*}{AUC}&   Indoor 0  &   Indoor 1  &   Indoor 2  &   Indoor 3  &   Outdoor 0  &   Outdoor 1  &   Outdoor 2  &   Outdoor 3  &   \multicolumn{3}{c}{Average}\\
&   (2 people)  &   (2 people)  &  (3 people)  &  (3 people)  &    (2 people)  &   (2 people)  &  (3 people)  &  (3 people) & (2 people) & (3 people) & (all) \\
\midrule
 {\bf Proposed}& 0.87 & 0.89 & \bf{0.88} & \bf{0.86} & 0.86 & 0.81 & 0.85  & 0.75  &  0.86 & \bf{0.84} & 0.84  \\
 {\bf Corr-only}& 0.86 & 0.90 & 0.83 & 0.80 & 0.85 & 0.75 & 0.81 & 0.71 &  0.84 & 0.79 & 0.80 \\
 {\bf Generic-only}& 0.85 &0.86 &0.86 &0.85 &0.81  &0.76  &0.81 &0.72 & 0.82 & 0.81 & 0.81 \\
 {\bf YKS}~\cite{Yonetani2015}& 0.88 & 0.89 & 0.75 & 0.72 & \bf{0.91} & \bf{0.88} & \bf{0.90}  & 0.70 & \bf{0.89} & 0.77 & 0.79  \\
 \midrule
 {\bf Proposed-ES}& \bf{0.91}& \bf{0.92} & \bf{0.88}& 0.84& 0.88 & 0.85 & 0.83 & \bf{0.80}& \bf{0.89} & \bf{0.84} & \bf{0.85}\\
 {\bf ES (Generic)}&0.80 &0.74 &0.81 &0.77 &0.80 &0.73 &0.75& 0.68 &0.77 & 0.75 &0.76 \\
 \midrule
 {\bf ZR}~\cite{Zhu2012}         &        0.60 &        0.61  &        0.64  &        0.58 &        0.65 &        0.69 &        0.67 &     0.62 &  0.64 & 0.63  & 0.63 \\
 {\bf VJ}~\cite{Viola2001}         &   0.66 & 0.73 & 0.66 & 0.62 & 0.75 & 0.77 & 0.73 & 0.62 &  0.73 & 0.66 & 0.67 \\
 {\bf LBPH}~\cite{Ahonen2006} & 0.50  & 0.50  & 0.51 & 0.54 & 0.57 & 0.55 & 0.55 & 0.50  &  0.53 & 0.52 & 0.52\\
  {\bf FisherFace}~\cite{Belhumeur1997} &0.50  & 0.50  & 0.51 & 0.54 & 0.50  & 0.52 & 0.50  & 0.51 & 0.51 & 0.51 &  0.51\\
\bottomrule \\
\toprule
\multirow{2}{*}{AP}&   Indoor 0  &   Indoor 1  &   Indoor 2  &   Indoor 3  &   Outdoor 0  &   Outdoor 1  &   Outdoor 2  &   Outdoor 3  &   \multicolumn{3}{c}{Average}\\
&   (2 people)  &   (2 people)  &  (3 people)  &  (3 people)  &    (2 people)  &   (2 people)  &  (3 people)  &  (3 people) & (2 people) & (3 people) & (all) \\
\midrule
 {\bf Proposed}& 0.52 & 0.50 & \bf{0.50} & \bf{0.49} & 0.49 & 0.45 & 0.48  & 0.42  &  0.49  &  \bf{0.47} & 0.48\\
 {\bf Corr-only}& 0.50 & 0.49 & \bf{0.50} & 0.48 & 0.47 & 0.41 & 0.45 & 0.39 &  0.47 & 0.45 & 0.46 \\
 {\bf Generic-only}& 0.52 &0.50 &\bf{0.50} &\bf{0.49} &0.48  &0.43  &0.48 &0.42 & 0.48 & 0.47 & 0.47 \\
 {\bf YKS}~\cite{Yonetani2015}& 0.49 & 0.48 & 0.40 & 0.38 & 0.49 & 0.50 & 0.51  & 0.38  &0.49 & 0.42 & 0.43  \\
 \midrule
 {\bf Proposed-ES}& 0.52& 0.51 & 0.46& 0.44& 0.50 & 0.48 & 0.46 & \bf{0.46}  & 0.50 & 0.45 & 0.47\\
 {\bf ES (Generic)}&0.43 &0.36 &0.40 &0.37 &0.44 &0.39 &0.42& 0.38 & 0.41 & 0.39  &0.39 \\
 \midrule
 {\bf ZR}~\cite{Zhu2012}         &        0.50 &        0.38  &        0.38  &        0.30 &        0.63 &        0.68 &        0.53 &     0.37 &  0.55 & 0.40 & 0.47 \\
 {\bf VJ}~\cite{Viola2001}         &   \bf{0.64} & \bf{0.60} & 0.32 & 0.27 & \bf{0.72} & \bf{0.75} & \bf{0.59} & 0.35 &  \bf{0.68} & 0.27 & \bf{0.53} \\
 {\bf LBPH}~\cite{Ahonen2006} & 0.51  & 0.49  & 0.46 & 0.35 & 0.57 & 0.55 & 0.55 & 0.43  &  0.53 & 0.45 & 0.49\\
  {\bf FisherFace}~\cite{Belhumeur1997} &0.27  & 0.0071  & 0.26 & 0.29 & 0.52  & 0.28 & 0.51  & 0.27 & 0.26 & 0.33 & 0.30\\
\bottomrule
\end{tabular}
}
\end{table*}

\begin{table}[t]
\centering
\caption{AUC and AP scores averaged over scenes in the CMU dataset. See Section~\ref{subsec:localization_experiment} for more details on each method. Note that no face recognition results were provided for Pool data because each participant was observed only in the single session.}
\label{tab:auc_cmu}
\scalebox{.9}{
\begin{tabular}{ccccc}
\toprule
\multirow{2}{*}{AUC}&   Pool &   Tennis &   Chat & \multirow{2}{*}{Average}\\
&   (2 people)  &   (3 people)  &  (4 people) &\\
\midrule
{\bf Proposed}&   \bf{0.86} &  \bf{0.83} &   0.83 & 0.83\\
{\bf Corr-only}&  0.81 &  0.77 &   0.77  & 0.78\\
{\bf Generic-only}&  \bf{0.86} &  \bf{0.83} &   0.81  & 0.82\\
{\bf YKS}~\cite{Yonetani2015}&   0.83 &  0.80 &   0.79 & 0.80 \\
\midrule
{\bf Proposed-ES}&   0.84 & 0.81 &   \bf{0.85} & \bf{0.84}\\
{\bf ES (Generic)}&  0.77 &  0.76 &   0.75  & 0.76\\
 \midrule
 {\bf ZR}~\cite{Zhu2012}        &   0.49  &           0.50  &   0.54  &0.53\\
 {\bf VJ}~\cite{Viola2001}       &   0.52 &           0.56 &   0.58 & 0.56\\
 {\bf LBPH}~\cite{Ahonen2006}&   - &           0.52 &   0.53 & 0.53 \\
 {\bf FisherFace}~\cite{Belhumeur1997}&   - &           0.53 &   0.53 & 0.53 \\
\bottomrule \\ 
\toprule
\multirow{2}{*}{AP}&   Pool &   Tennis &   Chat & \multirow{2}{*}{Average}\\
&   (2 people)  &   (3 people)  &  (4 people) &\\
\midrule
{\bf Proposed}&   \bf{0.50} &  \bf{0.45} &   0.47 & \bf{0.47}\\
{\bf Corr-only}&  \bf{0.50} &  \bf{0.45} &   0.45  & 0.45\\
{\bf Generic-only}&  \bf{0.50} &  \bf{0.45} &   0.47  & \bf{0.47}\\
{\bf YKS}~\cite{Yonetani2015}&   0.47 &  \bf{0.45} &   0.44 & 0.44 \\
\midrule
{\bf Proposed-ES}&   0.43 &  0.39 &   \bf{0.48} & 0.44\\
{\bf ES (Generic)}&  0.38 &  0.34 &   0.39  & 0.37\\
 \midrule
 {\bf ZR}~\cite{Zhu2012}        &   0.0025  &           0.0054  &   0.17  &0.12\\
 {\bf VJ}~\cite{Viola2001}       &   0.05 &           0.17 &   0.18 & 0.18\\
 {\bf LBPH}~\cite{Ahonen2006}&   - &           0.18 &   0.24 & 0.22 \\
 {\bf FisherFace}~\cite{Belhumeur1997}&   - &           0.33 &   0.30 & 0.31 \\
\bottomrule
\end{tabular}
}
\end{table}

\subsubsection*{Analysis on the efficient search}
{\bf Proposed-ES} worked comparably well with {\bf Proposed} and clearly outperforms {\bf ES (Generic)}. This is because data-driven generic targetness becomes high at \emph{any} target-like region and thus is not suitable to detect specific individuals. In addition to the choice hyper-parameters $P=25$ (the percentage of candidates to evaluate actual correlation-based targetness) and $K=32$ (the number of pieces for the piecewise constant approximation) used in {\bf Proposed-ES}, here we analyze how the proposed efficient search works on different settings of $P$ and $K$.

\begin{figure}[t]
\centering
\includegraphics[width=\linewidth]{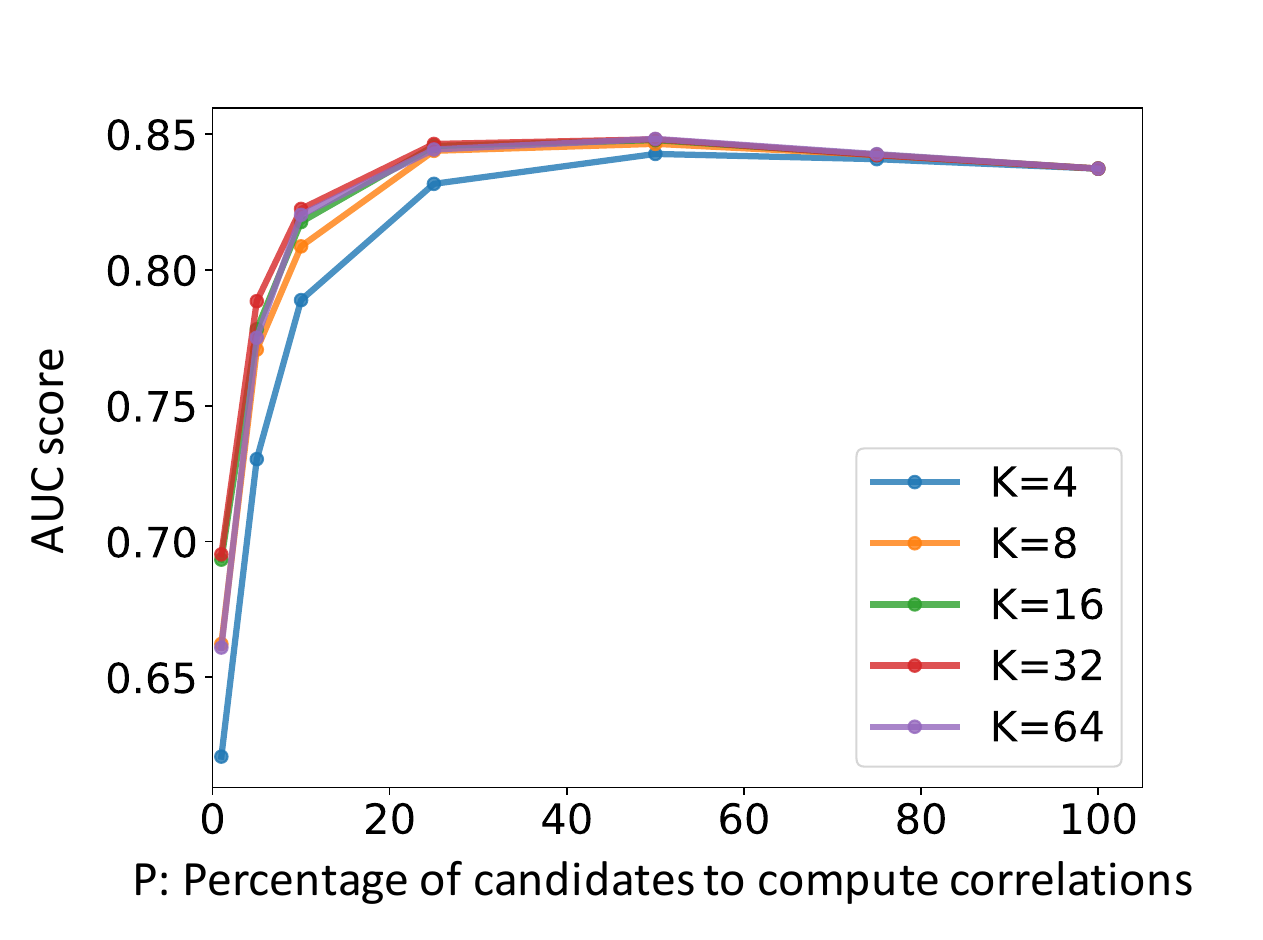}
\includegraphics[width=\linewidth]{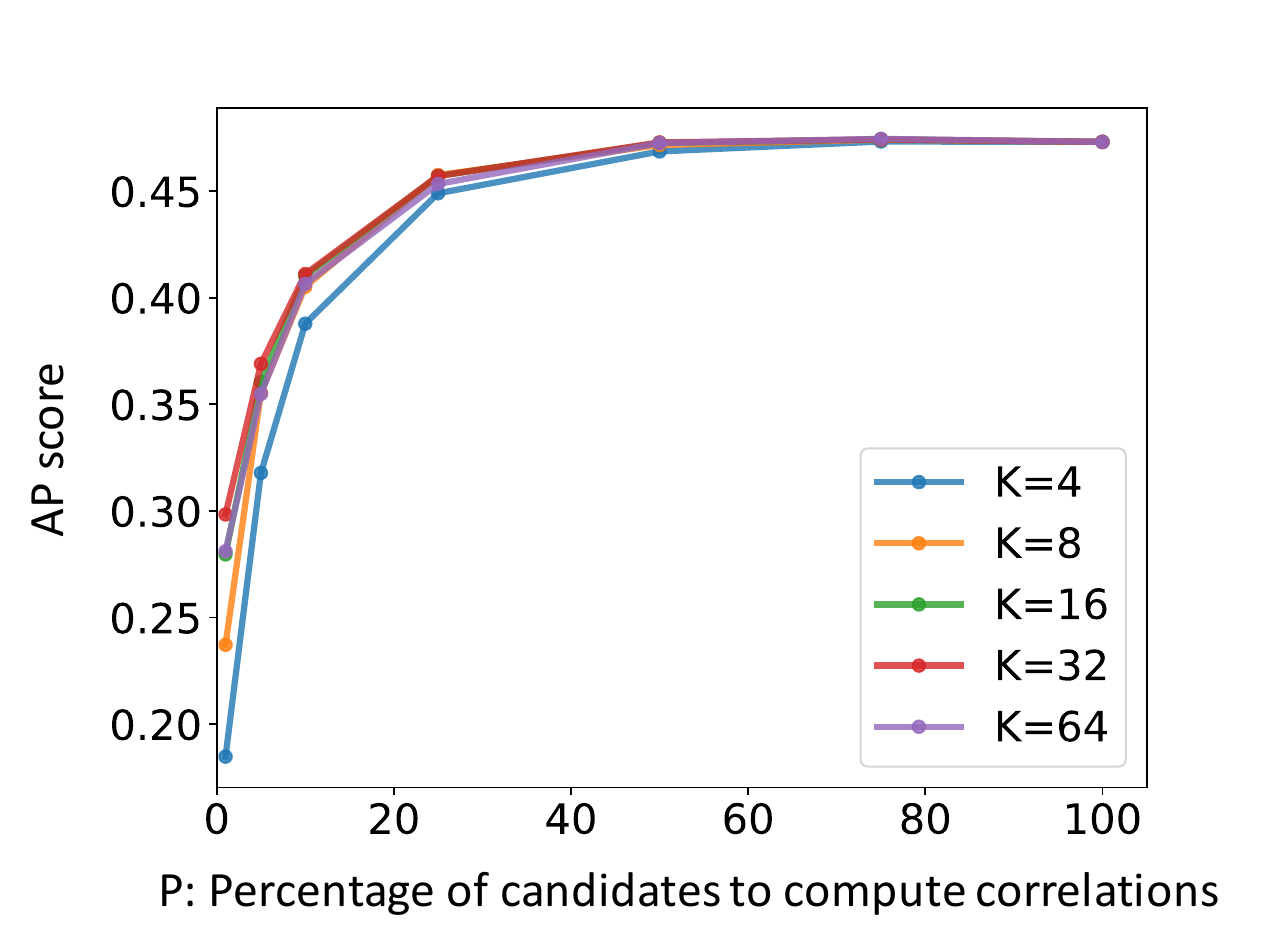}
\caption{Comparison of AUC and AP scores on different combinations of $P$: the percentages of candidates to evaluate actual correlation-based targetness and $K$: the numbers of pieces for the piecewise-constant approximation.}
\label{fig:performance_auc}
\end{figure}

Figure~\ref{fig:performance_auc} shows AUC and AP scores on the combinations of $P=1, 5, 10, 25, 50, 75, 100$ and $K=4, 8, 16, 32, 64$. Note that $P=100$ was equivalent to {\bf Proposed} in that all candidates were evaluated. The proper setting of $K$ depended on the frame rate to record videos as well as the maximum length of candidate trajectories $L\sub{max}$. Because we set $L\sub{max}=1024$ at 60 fps, piecewise constant approximations with $K=4$ could approximate about four-second motion patterns into one constant value, making it difficult to observe quick head motion. Nevertheless, our method maintained comparable AUC and AP scores when $P$ was large enough. The performance drastically dropped when $P<25$ and became comparable with baselines when $P=1$. When one needs to localize all instances in observer videos, $P$ should not be extremely small because the candidates may be incorrectly eliminated. For example, we found that the combination of $P=25$ and $K=32$ mostly worked fine in {\bf Proposed-ES} for a variety of scenes comprised in the two datasets.

\begin{figure*}[t]
\centering
\includegraphics[width=\linewidth]{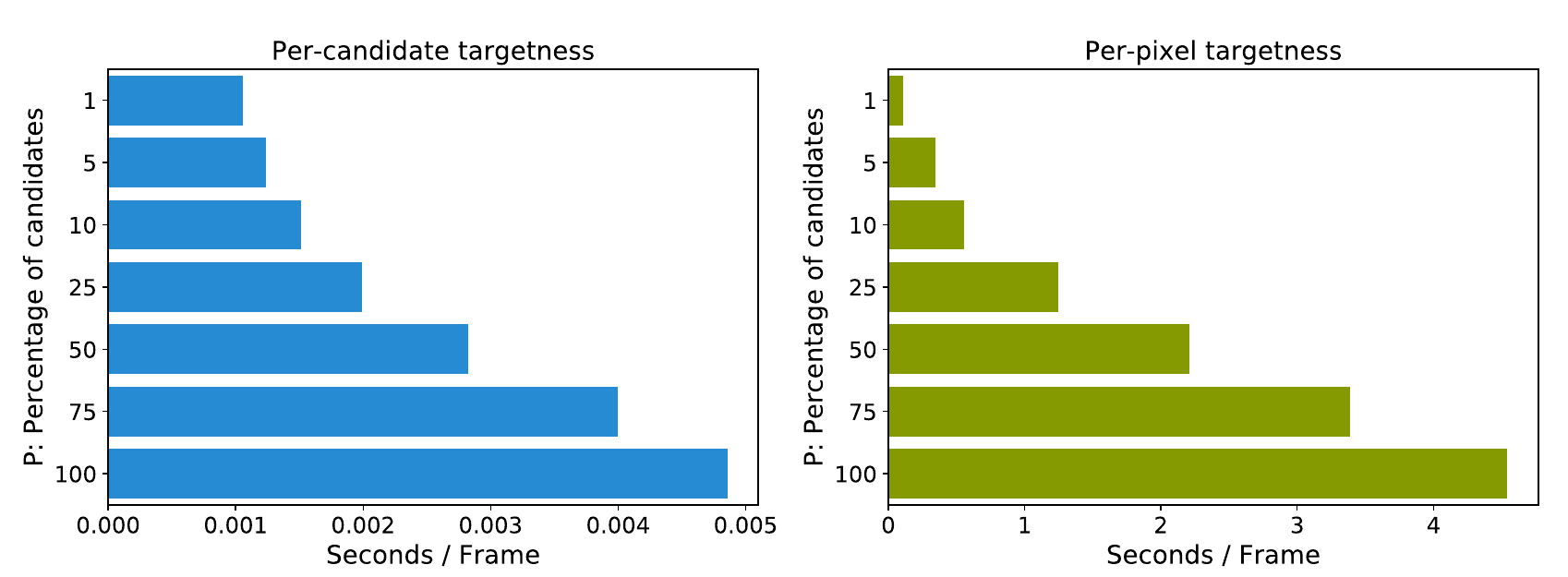}
\caption{Comparison of computation times on different percentages of candidates to evaluate actual correlation-based targetness. Left: computation times to evaluate per-candidate targetness. Right: computation times to evaluate per-pixel targetness maps.}
\label{fig:performance_speed}
\end{figure*}

Finally, we compared computation times\footnote{We implemented an overall procedure in Python and tested it on a single thread of MacPro with a 2.7 GHz 12-Core Intel Xeon E5.} on various $P$ while fixing $K$ to $K=32$ in Figure~\ref{fig:performance_speed}. Since generating queries and candidates requires a constant time regardless of $P$ (3 and 60 milliseconds per frame, respectively), here we focus on how computation times for per-candidate targetness evaluation and per-pixel targetness evaluation changed. {\bf Proposed-ES} reduced computation times by 69 percent compared with {\bf Proposed} on the per-candidate evaluation and by 88 percent on the per-pixel evaluation. Note that, when $P\leq 25$, {\bf Proposed-ES} improved computational efficiency over our prior approach {\bf YKS}~\cite{Yonetani2015} because the hierarchical supervoxel segmentation needed more than two seconds for each frame\footnote{We used LIBSVX with the default set of parameters provided with the code (http://www.cse.buffalo.edu/~jcorso/r/supervoxels/). See \cite{Yonetani2015} for more implementation details.}. \textcolor{black}{While the face detectors and recognizers used in our experiments were more efficient (\eg, about 25 milliseconds to detect faces with {\bf VJ}), their localization performances were quite limited as shown in Table~\ref{tab:auc_yks} and Table~\ref{tab:auc_cmu}.}

Accelerating per-candidate evaluation and per-pixel evaluation is both critical because they need to be applied for each combination of target and observer videos ($N\times M$ times for $N$ target and $M$ observer videos), while the procedures for generating queries and trajectory candidates needed to be done only once for each video ($N$ or $M$ times for the same setting). In addition, the acceleration on per-candidate evaluation is particularly beneficial for the tasks of target video retrieval and social group clustering which we will show in Section~\ref{subsec:affinity_experiment}; these tasks require only per-candidate targetness to compute an affinity between two first-person videos presented in Eq.~(\ref{eq:aff2}).

\subsubsection*{Target segmentation for privacy filtering}
\begin{figure}[t]
\centering
\includegraphics[width=\linewidth]{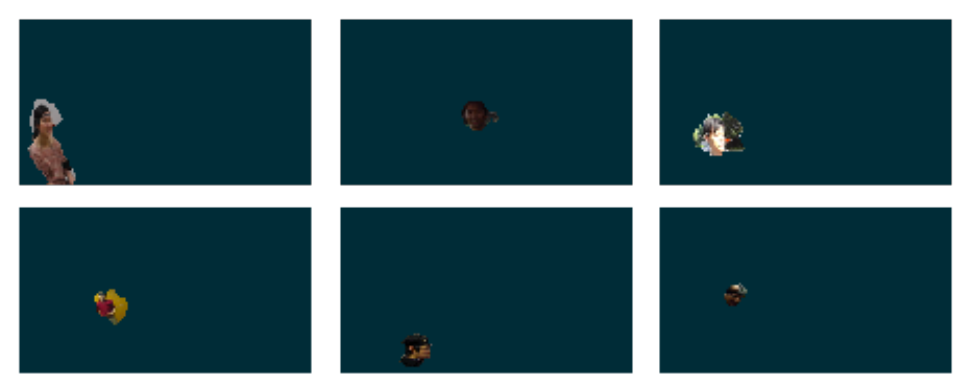}
\caption{Target segmentation with per-pixel targetness maps to preserve the privacy of people who accidentally came into the view of a camera.}
\label{fig:segmentation}
\end{figure}

Target localization results can be used to segment target regions in videos. In Figure~\ref{fig:segmentation} we ran GrabCut~\cite{Rother2004} on per-pixel targetness maps and found the region with the highest targetness score. Such segmentation is applicable for privacy filtering; by hiding regions other than target ones, we can preserve the privacy of people who accidentally came into the view of a camera.

\subsection{Application to Target Video Retrieval}
\label{subsec:affinity_experiment}
In this section, we show how our target search can be used for target video retrieval as a practical task. Consider a large collection of first-person videos shared publicly on a web service like YouTube. Given a first-person video recorded by a target individual as a query, we aim to retrieve videos including that target from the collection.

Taking into account the size of datasets evaluated in some related work using first-person videos for person identification (\eg, 50 videos used in~\cite{Ardeshir2016}), we combined our dataset (scenes {\bf Indoor 0 -- Indoor 3} and {\bf Outdoor 0 -- Outdoor 3}), the CMU dataset (scenes {\bf Pool, Tennis, Chat}), and the JPL interaction dataset~\cite{Ryoo2013}, which resulted in 113 videos in total. Note the JPL dataset comprises only observer videos and was used in the following experiment to see how our approach could identify people robustly where such irrelevant videos were present as a distractor.

We compared the affinity introduced in Eq.~(\ref{eq:aff2}) ({\bf Proposed}) to several other baselines: (1) {\bf B1}: using $P(a_{X^{(i)}}\mid \mathcal{V}_p)$ instead of $P(\mathcal{V}_p\mid a_{X^{(i)}})$ in Eq.~(\ref{eq:aff1}) and (2) {\bf B2}: using an asymmetric affinity $A(\mathcal{V}_p\mid\mathcal{V}_q)$ instead of $A(\mathcal{V}_p, \mathcal{V}_q)$ in Eq.~(\ref{eq:aff2}). For these methods, we set $P=25$ and $K=16$ to enable precise and efficient target search. In addition, we adopted some unsupervised scene descriptors for a baseline method based the observation that background objects were different across scenes in the datasets. Specifically, we used the GIST scene descriptor~\cite{Torralba2003}\footnote{We used the code available at http://lear.inrialpes.fr/software.} and deep features extracted using Places CNN~\cite{Zhou2014}\footnote{We used the pre-trained model Places205-GoogLeNet available at http://places.csail.mit.edu/downloadCNN.html. Responses in the last layer (\ie, posterior probabilities for each scene category) were used for the features.}. These features were extracted every 50 frames and were averaged over time to form a feature vector for each video.

\begin{figure*}[t]
\centering
\includegraphics[width=\linewidth]{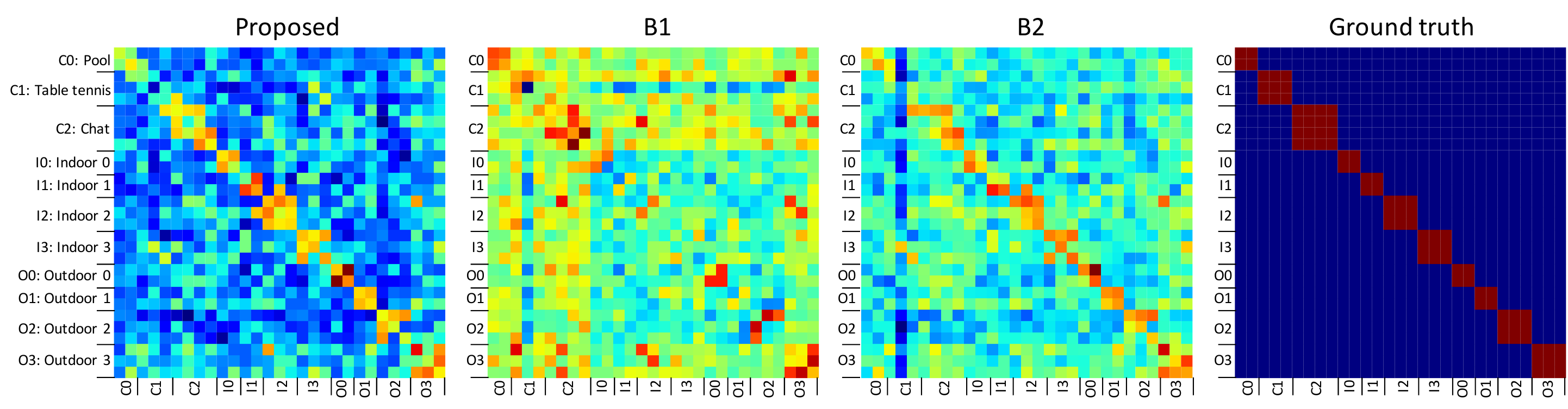}
\caption{Affinity matrices between first-person videos on our dataset and the CMU dataset.}
\label{fig:group}
\end{figure*}

Table~\ref{tab:search_evaluation} shows target video retrieval results. As the correct number of videos including target individuals was known, we evaluated the mean R-precision score~\cite{Manning2008} to compare methods quantitatively. Namely, we calculated the precision@$R$ for each target video, where $R$ is the number of videos including the target individual. Overall, we found that {\bf Proposed} performed the best. As shown in a part of affinity matrices in Figure~\ref{fig:group}, high affinity scores could be found in a block diagonal manner when using {\bf Proposed} and they were more similar to the ground truth affinity depicted in the right of the figure. On the other hand, the use of generic targetness in {\bf B1} made affinities more similar to each other making it unreliable for the target video retrieval task. We also found the performance degraded in {\bf B2}. This result indicates the importance of targetness of an opposite direction: how likely observers are included in the videos recorded by a target individual. 

\begin{table}[t]
\centering
\caption{Comparison of the mean R-precision on target video retrieval.}
\label{tab:search_evaluation}
\begin{tabular}{ccc}
\toprule
{\bf Proposed}& {\bf B1} & {\bf B2} \\
\midrule
{\bf 0.57} & 0.49 & 0.50\\
\midrule
{\bf GIST}~\cite{Torralba2003} & {\bf Places CNN}~\cite{Zhou2014} & \\
\midrule
0.26 & 0.30  &\\
\bottomrule
\end{tabular}
\end{table}

\subsection{Application to Social Group Clustering}Another possible application using our search is the task of social group clustering. Consider a scenario where a group of people equipped with a wearable camera are involved in certain group work like done in \cite{Kasahara2016,Park2012}. Given the recorded first-person videos, here we present a method to split them into several groups each of which shows the same social interaction. This method will be beneficial for analyzing multi-group interaction as it can automate annotations on who conversed with each other during the group work. To evaluate our approach on this task, we combined our dataset and the CMU dataset to generate a collection of first-person videos recorded by 29 people with 11 groups. As the number of groups is however unknown in practice, we adopted the affinity propagation algorithm~\cite{Frey2007} on the affinity matrices introduced in the previous section. Table~\ref{tab:group_evaluation} reported the precision, recall, and f-measure scores of group assignments as well as the estimated number of groups. We found that {\bf Proposed} again outperformed other methods in terms of clustering performances.

\begin{table}[t]
\centering
\caption{Performance Comparison on social group clustering. The ground-truth number of groups was 11.}
\label{tab:group_evaluation}
\begin{tabular}{ccccc}
\toprule
&   Precision &   Recall &   F-measure & \#groups\\
\midrule
{\bf Proposed}&   {\bf 0.81} &  {\bf 0.89} &  {\bf  0.84}  & {\bf 10} \\
{\bf  B1}&   0.60 &  0.78 &   0.66 & 8 \\
{\bf B2}&  0.59 &  0.65 &  0.60  & {\bf 10} \\
\midrule
{\bf GIST}~\cite{Torralba2003} &   0.24 & {\bf 0.89} &   0.36 & 3 \\
{\bf Places CNN}~\cite{Zhou2014} & 0.42& 0.74 & 0.49 & 6\\
\bottomrule
\end{tabular}
\end{table}
\section{Conclusions}
We introduced a novel correlation-based approach to the problem of self-search for first-person videos. Experimental results demonstrate that the proposed method was able to search and localize self instances robustly without the use of face recognition. Trajectory-based target candidates enabled accurate and efficient search over supervoxel candidates in our prior work~\cite{Yonetani2015}. The upperbound estimation of correlation scores was used to limit the number of candidates to apply a stable but slow evaluation of correlation-based targetness, making an overall search procedure more efficient while keeping high search performance.

The proposed method is beneficial for several practical applications such as monitoring specific individuals shown in videos of conversation scenes and analyzing social group interactions using wearable cameras. On the other hand, similar to some relevant work on person identification using first-person videos~\cite{Ardeshir2016,Fan2017,Hoshen2016}, our approach is not well suited for scenes where many people are moving in the same way (\ie, high correlation across multiple individuals)~\cite{Hoshen2016} and where individuals are heavily occluded and visible only for short periods of time (\ie, not enough signal)~\cite{Ardeshir2016,Fan2017}. Searching for people in these types of videos will require a richer set of features and will be an interesting direction for future work.

Extending the self-search to a larger repository of first-person videos will enable many novel applications. For example, searching for a group of people across first-person videos recorded at a variety of places around the world will illuminate their social activities, which have never been pursued by any visual surveillance. This application also raises new computer vision problems, such as social saliency prediction~\cite{Park2015} on a wide-spread area and group activity summarization for large-scale first-person videos.


%
%
%
%

\ifCLASSOPTIONcompsoc
  \section*{Acknowledgments}
\else
  \section*{Acknowledgment}
\fi

This research was supported by CREST JST and the Kayamori Foundation of Informational Science Advancement.

\ifCLASSOPTIONcaptionsoff
  \newpage
\fi



\bibliographystyle{IEEEtran}
%


%

\begin{IEEEbiography}[{\includegraphics[width=1in,height=1.25in,clip,keepaspectratio]{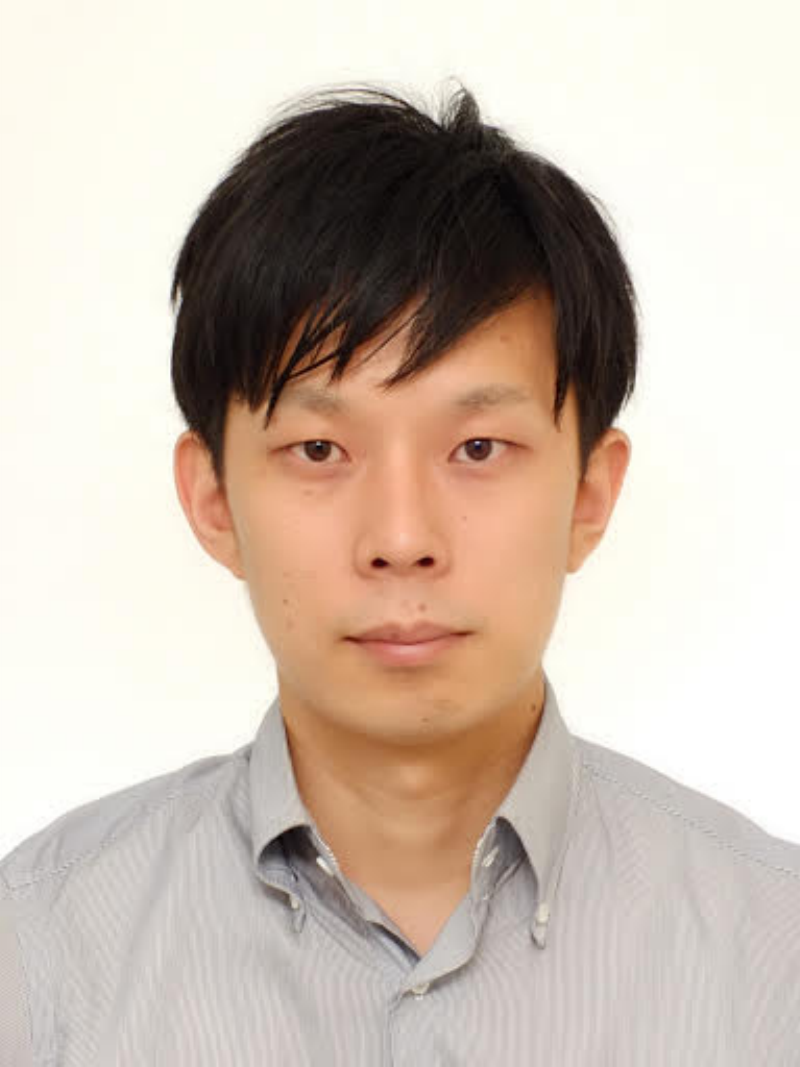}}]{Ryo Yonetani}
received his MS and PhD degrees in informatics from Kyoto University in 2011 and 2013. He is currently a 
research associate at Institute of Industrial Science, the University of Tokyo. His research interests include first-person vision, visual attention, and computer-human interaction.
\end{IEEEbiography}

\begin{IEEEbiography}[{\includegraphics[width=1in,height=1.25in,clip,keepaspectratio]{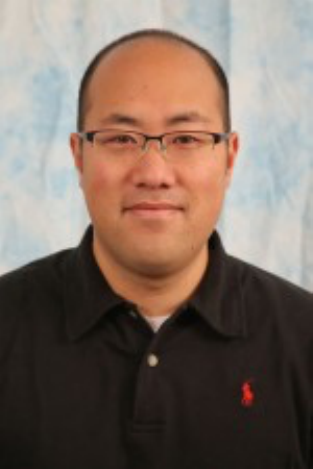}}]{Kris M. Kitani}
is an assistant research professor at the Robotics Institute at Carnegie Mellon University. He received his BS degree from the University of Southern California, and MS and PhD degree from the University of Tokyo. His research interests include human activity understanding, first-person vision and activity forecasting.
\end{IEEEbiography}


\begin{IEEEbiography}[{\includegraphics[width=1in,height=1.25in,clip,keepaspectratio]{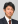}}]{Yoichi Sato}
is a professor at Institute of Industrial Science, the University of Tokyo. He received his B.S. degree from the University of Tokyo in 1990, and his MS and PhD degrees in robotics from School of Computer Science, Carnegie Mellon University in 1993 and 1997. His research interests include physics-based vision, reflectance analysis, first-person vision, and gaze sensing and analysis. He served/is serving in journal editorial roles including ones for IEEE Transactions on Pattern Analysis and Machine Intelligence, International Journal of Computer Vision, and Computer Vision and Image Understanding.
\end{IEEEbiography}




\end{document}